\documentclass{article}

\usepackage{microtype}
\usepackage{graphicx}
\usepackage{subfigure}
\usepackage{amsmath,amssymb,amsthm,amsfonts} 

\usepackage{booktabs} 

\usepackage{hyperref}

\usepackage[accepted]{icml2021}

\def\fig#1{fig.~\ref{#1}}
\def\eq#1{eq.~\ref{#1}}
\def\tab#1{table~\ref{#1}}
\def\sec#1{section~\ref{#1}}

\icmltitlerunning{Twin Neural Network Regression}

\begin{document}

\twocolumn[
\icmltitle{Twin Neural Network Regression}



\icmlsetsymbol{equal}{*}

\begin{icmlauthorlist}
\icmlauthor{Sebastian J. Wetzel}{1}
\icmlauthor{Kevin Ryczko}{2,3}
\icmlauthor{Roger G. Melko}{1,3,4}
\icmlauthor{Isaac Tamblyn}{2,3,5}
\end{icmlauthorlist}

\icmlaffiliation{1}{Perimeter Institute for Theoretical Physics, Canada}
\icmlaffiliation{2}{University of Ottawa, Canada}
\icmlaffiliation{3}{Vector Institute for Artificial Intelligence, Canada}
\icmlaffiliation{4}{University of Waterloo, Canada}
\icmlaffiliation{5}{National Research Council of Canada, Canada}

\icmlcorrespondingauthor{Sebastian J. Wetzel}{swetzel@perimeterinstitute.ca}

\icmlkeywords{Machine Learning, Artificial Neural Network, Regression, Ensemble Methods}

\vskip 0.3in
]



\printAffiliationsAndNotice{}  

\begin{abstract}


We introduce twin neural network (TNN) regression. This method predicts differences between the target values of two different data points rather than the targets themselves. The solution of a traditional regression problem is then obtained by averaging over an ensemble of all predicted differences between the targets of an unseen data point and all training data points. Whereas ensembles are normally costly to produce, TNN regression intrinsically creates an ensemble of predictions of twice the size of the training set while only training a single neural network. Since ensembles have been shown to be more accurate than single models this property naturally transfers to TNN regression. We show that TNNs are able to compete or yield more accurate predictions for different data sets, compared to other state-of-the-art methods. Furthermore, TNN regression is constrained by self-consistency conditions. We find that the violation of these conditions provides an estimate for the prediction uncertainty.

\end{abstract}

\section{Introduction}
Regression aims to solve one of two main classes of problems in supervised machine learning. It is the process of estimating the function that maps feature variables to an outcome variable. Regression can be applied to solve a wide range of problems. In everyday life, one may wish to predict the sales price of a house \cite{Park2015} or the number of calories in a meal \cite{Chokr2017}. In business and industry, it could be desirable to estimate stock market changes \cite{Patel2015} or the sales numbers of a certain product \cite{Sun2008}. Within the natural sciences, regression has been applied to a rich variety of problems, these include molecular formation energies \cite{rupp_prl_2012}, electronic charge distributions \cite{ryczko_pra_2019}, inter-atomic forces \cite{schnet}, electronic band-gaps \cite{Chandrasekaran2019}, and plasmonic response functions \cite{Malkiel2018}.

Indeed the wide range of the applications of regression makes it part of the standard curriculum across all of the quantitative domains (applied math, computer science, engineering, economics, the physical sciences, etc.).
There are many existing algorithms which solve the regression problem, including linear regression, Gaussian process regression, random forest regression, xgboost, and artificial neural networks, among others.

Regression problems require accurate and reliable solutions. Hence, in addition to the prediction itself, it is desirable to estimate the associated error or uncertainty \cite{Krzywinski2013, Ghahramani2015}. Such uncertainty signals can be used to decide when it is safe to let a model make decisions in the absence of expert supervision. For example, when a self-driving car experiences unfamiliar road conditions, model uncertainty can be used as a signal that it must alter its behavior \cite{J3016201806}. This could mean taking a different path, slowing down, or in the extreme, stopping until a human driver can take over. Similarly, in medical diagnostics, automated classification and analysis of diagnostic imaging can improve reliability and reduce costs \cite{McKinney2020}. However, such tools can only be trusted if they have the ability to gauge their own accuracy, and will only make recommendations when the expected prediction accuracy is above a safe threshold. Recent successes with surrogate models \cite{Ulissi2017, kasim2020billion} require an accurate estimate of model uncertainty. Similarly, active learning algorithms rely on an agent's self assessment ability; low confidence in a model can be used as a trigger for consulting the oracle \cite{Zhang2019, Zhong2020}. 
Methods to estimate the prediction uncertainty are often based on ensembles of predictions where the variation in output across models is used as a proxy for the uncertainty. This includes real ensembles, by combining the predictions of multiple models \cite{Hansen1990, Krogh1994}, or pseudo ensembles \cite{gal2015dropout,Bachman2014} that are obtained by perturbing certain parameters in the data or the model. While pseudo ensembles have the advantage of requiring no overhead in training time, real ensembles \cite{tran2020hydra} yield a much better prediction accuracy.
\begin{figure*}
    \centering
    \includegraphics[width=0.9\textwidth]{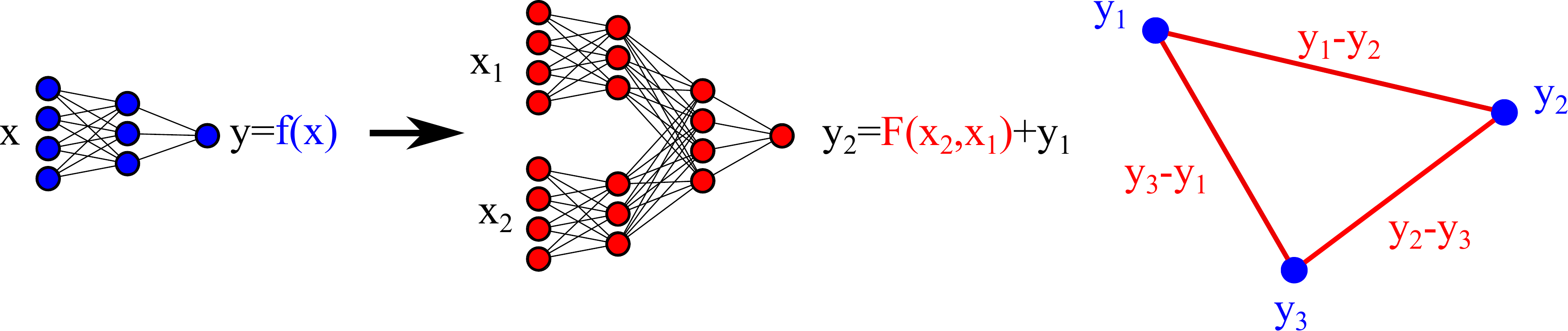}
    \caption{Reformulation of a regression problem: In the traditional case a neural network is trained to map an input $x$ to its target value $f(x)=y$. We reformulate the task to take two inputs $x_1$ and $x_2$ and train a twin neural network to predict the difference between the target values $F(x_2,x_1)=y_2-y_1$. Hence, this difference can be employed as an estimator for $y_2=F(x_2,x_1)+y_1$ given an anchor point $(x_1,y_1)$. }
    \label{fig:architecture}
\end{figure*}
In this paper we present a regression algorithm based on twin neural networks (TNN) (historically known as Siamese neural networks) that leverage sample-to-sample comparisons while making their predictions. Twin neural networks were introduced in \cite{Baldi1993} for finger print identification and in \cite{siamese} for hand written signature identification. They are popularly used in few shot learning \cite{Koch2015} or facial recognition applications \cite{Taigman2014}.

Our method naturally produces an ensemble of predictions when it compares the predictions between a new unseen data point and all training data points. It combines the strengths of real and pseudo ensembles. On one hand, it creates a large number of predictions (twice size of the training data set) at little additional cost in training time compared to a traditional neural network. On the other hand, as a real ensemble it significantly increases the prediction accuracy.

TNN regression also provides estimates of model uncertainty by construction of a network topology which allows for self-consistency conditions. A violated check on these conditions can be interpreted as a decrease in prediction accuracy. These checks include, among others, the standard deviation associated with the prediction.

We first describe the approach, demonstrate its performance on well known data sets, and finally examine its self-consistency conditions.

The main contributions of the paper of the paper are the development of a new regression algorithm, which produces a large ensemble of predictions at low cost. This algorithm is shown to be competitive and outperforms state-of-the-art on many datasets. Further, self-consistency conditions can be used to estimate the prediction error.



\section{Prior Work}
Twin (or Siamese) neural network (TNN) regression (\fig{fig:architecture}) yields high accuracy and uncertainty estimates based on the intrinsic ensemble it leverages.

TNNs were originally developed as an approach for the identification of fingerprints \cite{Baldi1993} and handwritten signature verification \cite{siamese}. More recently, when coupled with deep convolutional architectures \cite{LeCun1989}, TNNs have been used for facial recognition \cite{Taigman2014}, few shot learning\cite{Koch2015}, and object tracking \cite{Bertinetto2016}. The idea of pairwise similarity has also been shown to be an approach for unlabeled classification \cite{bao2018classification}. TNNs have previously been used in the regressive task of extracting a camera pose from images \cite{doumanoglou2016siamese}. Other uses of TNNs with images and regression have been focused in the medical domain as an estimator for disease severity \cite{Li2020}. The ability of TNN to extract similarities from data can also be used to determine conservation laws in physics \cite{wetzel2020discovering}.



Uncertainty assessments for linear regression are well established. Similarly, for stochastic processes, there are standard techniques which can estimate the uncertainty for a model. Gaussian processes (GP) \cite{Bartok2010, Koistinen2016, Simm2018, Proppe2019} naturally provide an estimate of uncertainty, but the cost of training grows quickly with the number of training samples; GP are impractical for large data-sets, although there has been recent progress in this direction \cite{Liu2019}. GP also are unable to easily incorporate new data; it is necessary to retrain the model fully for each new data point or observation, making online learning very costly.




Conversely, there is not yet a single, established protocol for quantifying error for (deep) neural networks, particularly for the case of regression. An early and straightforward approach to uncertainty estimation is the use of ensembles. 

Ensemble methods are commonly used in regression (and classification) tasks as a means of improving the accuracy of prediction \cite{naftaly1997} and solving the bias-variance tradeoff \cite{pattern_recog_and_ml} by combining the predictions of different models. Ensembles \cite{Hansen1990, Krogh1994} can be produced in different ways such as repetition of training while changing training-validation splits or even sampling intermediate models along the training trajectory \cite{Swann1998, Xie2013, huang2017snapshot}. Sampling along a training trajectory is efficient in that it generates approximately 5 ensemble members in the time it takes to train one traditional neural network. 

The disagreement between models in an ensemble can be used as a signal for confidence among the set. Recent work has highlighted the problems associated with ensembles as a method for uncertainty estimation \cite{ashukha2020pitfalls,lakshminarayanan2017simple,Cortes-Ciriano2019}. There is less theoretical justification for the reliability of errors from such approaches compared to GP. Intuitively it makes sense that a mismatch of models suggests the output cannot be trusted, but it is less clear that the magnitude of this mismatch can be assigned to a particular value of uncertainty.

An alternative method to a real ensemble is to sample the output of a network subject to perturbations, for example through the introduction of errors. This can be regarded as a form of pseudo ensemble \cite{Bachman2014}. MC dropout \cite{gal2015dropout} samples a network where nodes are randomly deactivated. In this case, error signals come from the fact that the collective response needed to overcome de-activations depends on the distribution of data itself. If data far away from the training regime is used, regression becomes unreliable (this is a well known effect). It also means that the learned corrections (e.g. the response of the remaining nodes) become less reliable - hence the variance observed when sampling the network output increases. MC dropout is successful because the topology is constructed implicitly to signal whether the output can be trusted.

Other methods for estimating model uncertainty and error include discriminant analysis \cite{Morais2019}, resampling \cite{Musil2019}, scoring rules \cite{Gneiting2007, Dawid2014}, and domain specific metrics \cite{Peterson2017, Liu2018, Tran2020}. Finally, it was recently shown that the projection of data into the latent space of a model can be used as a proxy for uncertainty; a close distance to one of the training points is correlated with lower error \cite{Janet2019}.

\section{Twin Neural Network Regression}
\begin{table*}
\centering
  
  \begin{tabular}{l lll lll}
       & \multicolumn{6}{l}{Common Data}           \\
    \cmidrule(r){2-7}
  &BH&CS&EE&YH&WN&BC\\
  RF &$4.24\!\pm\!0.29$&$8.23\!\pm\!0.24$&$2.22\!\pm\!0.08$&$2.95\!\pm\!0.46$&$0.64\!\pm\!0.02$&$\mathbf{0.71\!\pm\!0.03}$\\
  
  XGB &$2.93\!\pm\!0.18$&$4.37\!\pm\!0.19$&$1.17\!\pm\!0.04$&$\mathbf{0.42\!\pm\!0.06}$&$\mathbf{0.61\!\pm\!0.01}$&$\mathbf{0.70\!\pm\!0.03}$\\

  ANN &$3.09\!\pm\!0.14$&$5.37\!\pm\!0.17$&$0.98\!\pm\!0.03$&$0.52\!\pm\!0.07$&$0.64\!\pm\!0.01$&$0.76\!\pm\!0.02$\\

  ANNE &$3.43\!\pm\!0.32$&$5.14\!\pm\!0.21$&$0.89\!\pm\!0.04$&$\mathbf{0.43\!\pm\!0.05}$&$\mathbf{0.62\!\pm\!0.01}$&$\mathbf{0.72\!\pm\!0.03}$\\

  MCD &$2.95\!\pm\!0.15$&$6.07\!\pm\!0.21$&$2.96\!\pm\!0.12$&$1.42\!\pm\!0.18$&$0.68\!\pm\!0.01$&$\mathbf{0.72\!\pm\!0.03}$\\

  TNN &$\mathbf{2.55\!\pm\!0.10}$&$4.19\!\pm\!0.25$&$0.52\!\pm\!0.02$&$0.49\!\pm\!0.07$&$\mathbf{0.62\!\pm\!0.01}$&$0.83\!\pm\!0.03$\\

  TNNE &$\mathbf{2.61\!\pm\!0.20}$&$\mathbf{3.88\!\pm\!0.22}$&$\mathbf{0.46\!\pm\!0.02}$&$\mathbf{0.37\!\pm\!0.06}$&$0.63\!\pm\!0.01$&$\mathbf{0.72\!\pm\!0.02}$\\ \\
  \end{tabular}
  \begin{tabular}{l llllll}
     & \multicolumn{3}{l}{Science Data}    & &  & \multicolumn{1}{l}{Image Data}           \\
    \cmidrule(r){2-4}\cmidrule(r){7-7}
  &RP&RCL&WSB&&&ISING\\
  RF &$0.604\!\pm\!0.013$&$0.288\!\pm\!0.004$&$0.141\!\pm\!0.011$&&RF&$0.601\!\pm\!0.003$\\
  
  XGB &$0.229\!\pm\!0.005$&$0.124\!\pm\!0.002$&$0.071\!\pm\!0.006$&&XGB&$0.144\!\pm\!0.003$\\

  ANN &$0.050\!\pm\!0.002$&$0.019\!\pm\!0.000$&$0.047\!\pm\!0.004$&&CNN&$0.050\!\pm\!0.001$\\

  ANNE &$0.032\!\pm\!0.002$&$0.016\!\pm\!0.001$&$0.031\!\pm\!0.002$&&CNNE&$0.044\!\pm\!0.001$\\

  MCD &$0.086\!\pm\!0.002$&$0.033\!\pm\!0.001$&$0.042\!\pm\!0.003$&&CMCD&$0.052\!\pm\!0.001$\\

  TNN &$0.022\!\pm\!0.001$&$0.017\!\pm\!0.000$&$\mathbf{0.020\!\pm\!0.001}$&&CTNN&$0.035\!\pm\!0.001$\\
  

  TNNE&$\mathbf{0.016\!\pm\!0.001}$&$\mathbf{0.014\!\pm\!0.001}$ &$0.022\!\pm\!0.002$ &&     CTNNE&$\mathbf{0.030\!\pm\!0.001}$\\
  
  \end{tabular}
  
  \caption{Best estimates for root mean square errors (RMSEs) of different algoritms on the test sets belonging to different data sets. The Lowest RMSEs are in bold for clarity. Our confidence on the RMSEs is determined by their standard error. Data sets: Boston housing (BH), concrete strength (CS), energy efficiency (EE), yacht hydrodynamics (YH), red wine quality (WN), Bio Conservation (BC), random polynomial (RP), RCL circuit (RCL), Wheatstone bridge (WSB) and the Ising Model (ISING). Algorithms: Random Forests (RF), xgboost (XGB), Neural Networks (ANN), Monte-Carlo Dropout networks (MCD), Twin Neural Networks (TNN) and ensembles (E) or convolutional variants (C).
  }
  \label{tab:results}
\end{table*}
\subsection{Reformulation of Regression}
In a regression problem we are given a training set of $n$ data points $X^{train}=(x_1^{train},...,x_n^{train})$ and target values $Y^{train}=(y_1^{train},...,y_n^{train})$. The task is to find a function $f$ such that $f(x_i)= y_i$. Further, we require that the function $f$ does generalize to unseen data $X^{test}$ with labels $Y^{test}$. In the following we reformulate this regression problem. Given a pair of data points $(x_i^{train},x_j^{train})$ we train a neural network (\fig{fig:architecture}) to find a function $F$ to predict the difference 
\begin{align}
    F(x_i,x_j)= y_i-y_j \quad .    
\end{align}
This neural network can then be used as a solution for the original regression problem $y_i= F(x_i,x_j)+y_j$. In this setting we call $(x_j,y_j)$ the {\em anchor} for the prediction of $y_i$. This relation can be evaluated on every training data point $x_j^{train}$ such that the best estimate for the solution of the regression problem is obtained by averaging
\begin{align}
    y_i^{pred}&=  \frac{1}{n}\sum_{j=1}^n F(x_i,x_j^{train})+y_j^{train}\\ 
    &= \frac{1}{n}\sum_{j=1}^n \frac{1}{2}F(x_i,x_j^{train})-\frac{1}{2}F(x_j^{train},x_i)+y_j^{train}  \, .\label{eq01}
\end{align}
The first advantage of the reformulation is that it creates an ensemble of twice the size of the training set of predictions of differences $y_i-y_j$ for a single prediction of $y_i$. While ensembles are in general costly to produce, the TNN regression intrinsically yields a very large ensemble at little additional training cost.

In general, we do not expect the ensemble diversity of the TNN regression to be similar to traditional ensembles. The reason is that the prediction of a regression target $y_i$ is based on different predictions for differences $y_i-y_j$ due to multiple anchor points. This allows us to combine the TNN regression with any traditional ensemble method to achieve an even more accurate prediction.

Each prediciton of $y_i$ is made from a finite range of differences $y_i-y_j$ and the anchor data points differ by more than an infinitesimal perturbation. Hence, the TNN regression ensemble is not just a pseudo ensemble that is obtained by small perturbations of the model weights. 

The intrinsic ensembles of the TNNs are not conventional ensembles. Like k-nearest neighbor regression or support vector regression the prediction is formed by leveraging comparison between a new unseen datapoint and several support vectors or nearest neighbors belonging to the training set. However, in contrast to these algorithms, TNN regression can be seen from a single neural network perspective with weight diversity. Let's consider the prediction $y_i=F(x_i,x_j)+y_j$, we can consider $x_i$ as input to a traditional model to predict $y_i$, while $x_j$ can be understood as auxiliary parameters which influence the weights. The offset $y_j$ can be seen as changing the bias of the output layer.

In principle a reliable estimation of a regression target through \eq{eq01} might be prone to a larger error compared to a traditional estimation. The reason for this is that the TNN regression must make predictions on two data points and thus also accumulates the errors of both inputs. However, since we average over the whole training set the errors on the anchor data points are uncorrelated and thus average out leading to a suppression by a factor of $1/\sqrt{2n}$, where $n$ is the training set size.

\subsection{Self-Consistency Conditions}
The twin neural network predicts the difference between two regression targets. For an accurate prediction this requires the satisfaction of many self-consistency conditions, see \fig{fig:architecture}. Any closed loop of difference predictions $F(x_i,x_j)=y_i-y_j$ sums up to zero.
Any violation of such a condition is an indication of an inaccurate prediction. In principle there are several ways to harness this self-consistency condition for practical use. First, it can be used to estimate the magnitude the prediction error. Second, it could be utilized to force the neural network to satisfy these conditions during training. Finally, it can enable one to use the predictions on previous test data as anchor points for new predictions.

The smallest loop only contains two data points $x_i,x_j$ for which an accurate TNN needs to satisfy
\begin{align}
0=y_i-y_j+y_j-y_i=F(x_i,x_j)+F(x_j,x_i)\label{eq:sc2}
\end{align}
While training, in each batch we include both pairs $(x_i,x_j)$ and its mirror copy $(x_j,x_i)$ to enforce the satisfaction of this condition while training. The predictions on any three data $x_i,x_j,x_k$ points should satisfy
\begin{align}
0=F(x_i,x_j)+F(x_j,x_k)+F(x_k,x_i)
\end{align}
For $x_i,x_j\in X^{train}$ the target values $y_i,y_j$ are known. Thus, this condition becomes equivalent to the statement that the prediction of $y_k$ must be the same on any two different anchor points $(x_i,y_i)$ and $(x_j,y_j)$.
\begin{align}
y_k=F(x_k,x_i)+y_i=F(x_k,x_j)+y_j\label{eq:sc3}
\end{align}
This condition is trivially enforced during training. We examine the relation of magnitudes of the violations of these conditions and the prediction error in \sec{sec:pred_error}. To this end we employ the ensemble of predictions and calculate the standard deviation corresponding to \eq{eq:sc2} and \eq{eq:sc3} and find a distinct correlation with the out-of-domain prediction error .

\subsection{Twin Neural Network Architecture}
The reformulation of the regression problem does not require a solution by artificial neural networks. However, neural networks scale favorably with the number of input features in the data set. We employ the same neural network for all data sets. Our TNN takes a pair of inputs $(x_i,x_j)$, where each input is connected to the fully connected neural network with two hidden layers and a single output neuron. Each hidden layer consists of 64 neurons each with a \emph{relu} activation function. On data sets containing hierarchical structures, such as image data sets or audio recordings, it is helpful to include shared layers that only act on one element of the input pair. This is commonly used in few shot learning in image recognition \cite{siamese}. We optimize a common architecture that works well for all data sets considered in this work. We examined different regularization methods like dropout and $L_2$ regularization and found that in some cases a small $L_2$ penalty improves the results. More details can be found in the supplementary materials. The improvement was not statistically significant or uniform among different splits of the data, which is why our main results omit any regularization. The training objective is to minimize the \emph{mean squared error} on the training set. For this purpose we employ standard gradient descent methods \emph{adadelta} (and \emph{rmsprop}) to minimize the loss on a batch of 16 pairs at each iteration. We stop the training if the validation loss stops decreasing.

The single feed forward neural network (ANN) that we employ for our comparisons has a similar architecture as the TNN. This means it has the same hidden layers and we examined the same hyperparameters. The convolutional neural networks are build on this architecture with the first two dense layers exchanged by convolutional layers of 64 neurons and filter size 3.
The neural networks are robust with respect to changing the specific architectures in the sense that the results do not change significantly. The neural networks were implemented using keras \cite{Chollet2015} and tensorflow \cite{Tensorflow2015}.
\begin{figure}[ht]
\vskip 0.2in
\begin{center}
\centerline{\includegraphics[width=\columnwidth]{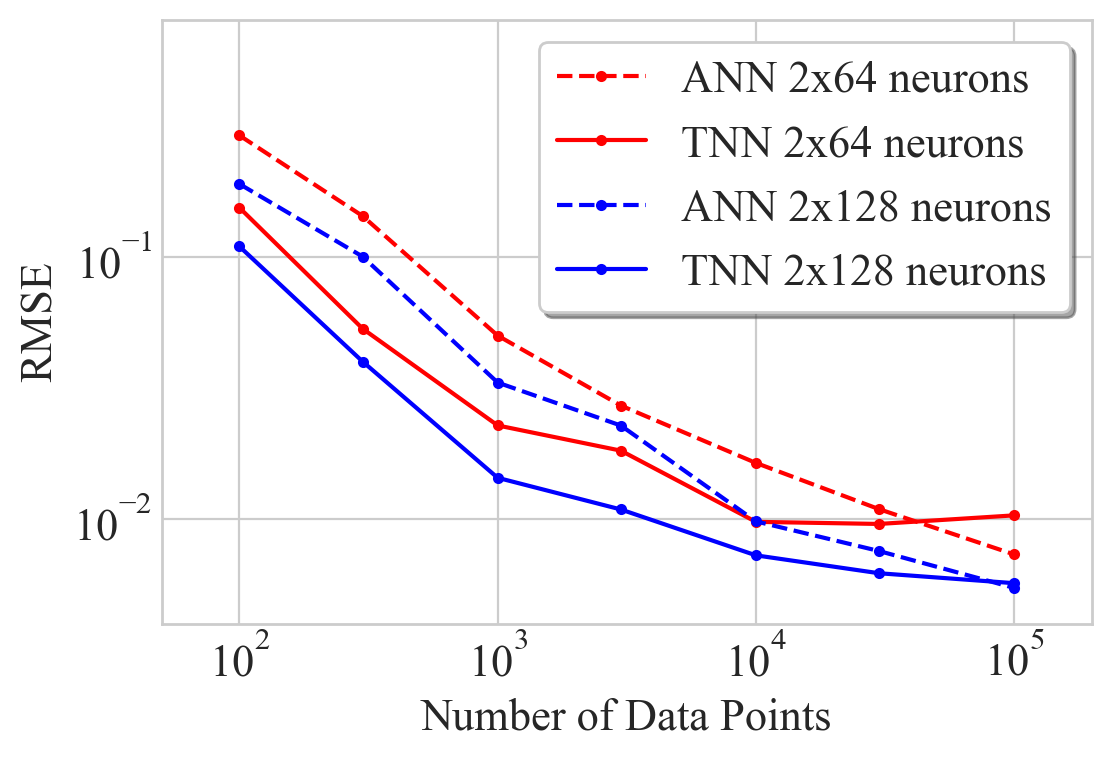}}
\caption{Traditional ANN and TNN regression applied to the random polynomial function (RP). Standard architectures ($2\times 64$ hidden neurons) and optimized architectures ($2\times 128$ hidden neurons) are trained with different training data sets containing $n=100\dots 100000$ data points. Larger training sets reduce RMSE, TNN regression beats ANN regression for $n < 60000\dots 100000 $.}
\label{fig:data_rmse}
\end{center}
\vskip -0.2in
\end{figure}

\begin{figure*}
  \centering
  \includegraphics[width=0.9\textwidth]{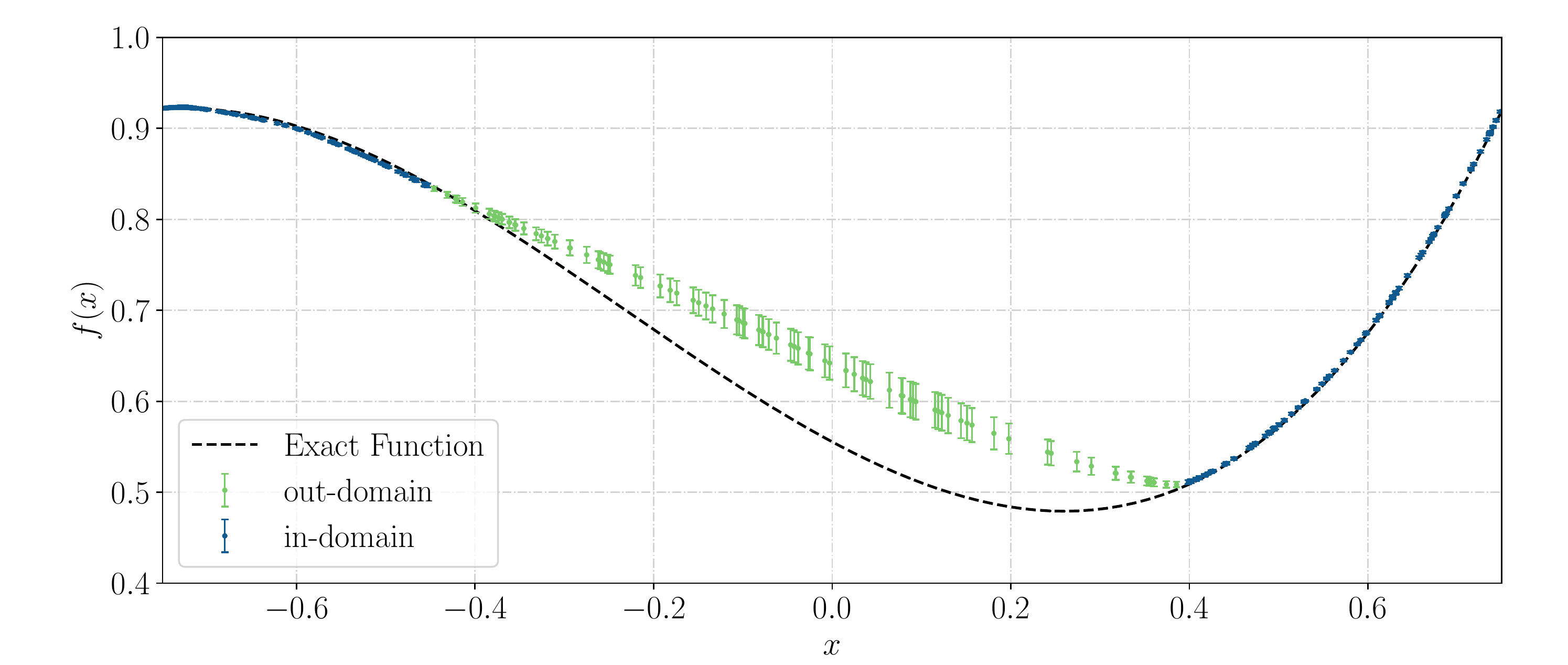}
  \caption{TNN regression applied to a simple function to demonstrate its behavior outside of the training domain (interpolation in this case). For intervals within the training domain, the TNN is able to accurately reproduce the original  function (black dotted lines) well. Over this interval the model has low uncertainty measured by the standard deviation of the TNN ensemble prediction. This is equivalent to the satisfaction of the self-consistency condition \eq{eq:sc3}. Conversely, within the interval which was obscured during training, the performance of the model is poor. The corresponding higher uncertainty or violation of the self-consistency conditions is a signal that the model is performing poorly.}
  \label{fig:1dpoly}
\end{figure*}

\subsection{Data Preparation}
\label{sec:data}
We examine the performance of TNN regression on different data sets: Boston housing (BH), concrete strength (CS), energy efficiency (EE), yacht hydrodynamics (YH), red wine quality (WN), Bio Conservation (BC),random polynomial (RP), RCL circuit (RCL), Wheatstone bridge (WSB) and the Ising Model (ISING). The common data sets can be found online at \cite{datasets}. The science datasets are simulations of mathematical equations and physical systems. RP is a polynomial of degree two of five input features and random coefficients. RCL is a simulation of the electric current in an RCL circuit and WSB a simulation of the Wheatstone bridge voltage. ISING, a spin system on a lattice of size $20\times20$ and the corresponding energies are used to demonstrate an image regression problem. More details can be found in the supplementary matierial.

All data is split into 90\% training, 5\% validation and 5\% test data. Each run is performed on a randomly chosen different split of the data. We normalize and center the input features to a range of $[-1,1]$ based on the training data.
In the context of uncertainty estimation we further divide all data based on their regression targets $y$. In this manner we choose to exclude a certain range from the training data. Hence, we can examine the performance of the neural networks outside of the training domain.

While the data can be fed into a standard ANN in a straightforward manner, one must be careful in the preparation of the TNN data. Starting with a training data set of $n$ data points we can create $n^2$ different pairs of training data pairs for the TNN. Hence, the TNN has the advantage of having access to a much larger training set than the ANN. However, in the case of a large number of training examples, it does not make sense to store all pairs due to memory constraints. That is why we train on a generator which generates all possible pairs batchwise before reshuffling.

\section{Experiments}
\subsection{Prediction Accuracy}

We train random forests(RF), xgboost(XGB), traditional single neural networks ANNs, TNNs and ensembles to solve various regression tasks outlined in \sec{sec:data}. The hyperparameters of RF and XGB are optimized for each dataset via cross-validation. The performance on these data sets is measured by the \emph{root mean square error} (RMSE) which is shown in \tab{tab:results}. Each result is averaged over 20 different random splits of the data; in this manner we find the best estimate of the RMSE and the according standard error. We also produce explicit ensembles (E) by training the according ANN and TNN 20 times. This means each RMSE in the table requires training 20 times or 400 times for the ensembles (E).

In \tab{tab:results} we see that TNNs outperform single ANNs and even ensembles of ANNs on all data sets except BC, we assume this is an statistical outlier. Creating an ensemble of TNNs increases the performance even further. The significance of the difference in performance is clearly supported by the comparably small standard error. On discrete data especially YH, WN and BC, XGB slightly outperforms ANNs at a much shorter training time. However, TNNs are able to compete. In science data sets, due to the continous variables, neural network based methods beat tree based methods significantly. On image data convolutional neural networks outperform RF and XGB even though ISING is discrete. The general trend is that TNNs outperform ANNs. However, the TNN takes 15 to 40 times longer to converge compared to single ANNs. In the case of Boston Housing this would translate to approximately 1000 ensemble members in 40$\times$ the time, or 25 ensemble members per time equivalent. 

Convolutional TNN architectures as employed for the Ising Model energy predictions are nontrivial, the present result presents a proof of concept in \cite{Ryczko2020} where it is further explained and extended.

The out-performance of TNNR over other regression algorithms is based on exploiting the pairing of data points to create a huge training data set and a large ensemble during inference time. If the training set is very large, the number of pairs increases quadratically to a point where the TNN will in practice converge to a minimum before observing all possible pairs. At that point the TNN begins to lose its advantages in terms of prediction accuracy. A visualization of this fact can be seen in \fig{fig:data_rmse}. In this figure different ANN and TNN architectures are applied to the RP data set. One may observe the performance improvement in terms of lower RMSE of all neural networks when increasing the number of training data points. The TNN achieves a lower RMSE than the ANN in small and medium sized data sets.The TNN finds a plateau sooner than the ANN such that  both algorithms perform similarly well in a regime of betweem $60000$ and $100000$ data points. We infer that if the training set is sufficiently large an ANN starts to compete with or outperform the TNN.

\begin{table*}
\centering

  \centering
  \begin{tabular}{lllllll}
    \toprule
    &\multicolumn{3}{c}{ANN MC Dropout}   &   \multicolumn{3}{c}{TNN}             \\
         & RMSE$_{train}$     &  RMSE$_{test_{in}}$ &  RMSE$_{test_{out}}$ & RMSE$_{train}$     &  RMSE$_{test_{in}}$  &  RMSE$_{test_{out}}$\\
    \midrule
    
    BH& 2.643  & 4.429 & 11.540 & 1.137 & 4.468   & 11.729\\
    CS    & 5.505 & 6.639 & 10.199 & 3.096 &    4.408  & 5.000\\
    EE     &   3.373     & 3.737 & 3.689 & 0.741 & 1.329 & 2.187\\
    RP & 0.046  & 0.075 & 0.520 & 0.015 & 0.022 & 0.207  \\

    \bottomrule
  \end{tabular}
  
    \caption{Comparison of ANN Dropout ensembles and TNN ensembles with division of testing data into $test_{in}$ which is on the training manifold and $test_{out}$ which is outside the training manifold in the context of extrapolation.
  }
  \label{tab:manifold_results}
\end{table*}
\subsection{Prediction Uncertainty Estimation}
\label{sec:pred_error}

Equipped with a regression method that yields a huge ensemble of predictions constrained by self-consistency conditions we examine reasonable proxies for the prediction error. 

In that sense we must distinguish between in-domain prediction uncertainty and out-of-domain prediction uncertainty. The in-domain denotes data from the same manifold as the training data. However, if the test manifold differs from the training manifold one is concerned with the out-of-domain data. This is the case in interpolation or extrapolation, or if certain features in the test data exhibit a different correlation as in the training data. 

Consider for example the polynomial function of one variable (\fig{fig:1dpoly}) as an example of an out-of-domain interpolation problem. The TNN is very accurate on the test set as long as one stays on the training manifold. As soon as the test data leaves the training manifold the prediction becomes significantly worse. While the prediction resembles a line connecting the endpoints of the predictions on the training manifold it fails to capture the true function. Since the difference between the true function and the regression result is consistently larger than the standard deviation, we further conclude that the standard deviation of the TNN regression result cannot be used as a direct estimate for the prediction uncertainty.

In the following we examine the possibility of a meaningful estimation of the prediction uncertainty on the BH, CS, EE and RP data sets. Before training we separate $25 \%$ of the data as out-of-domain test set $test_{out}$ dependent on a threshold on the regression target $y$ to simulate an extrapolation problem. We use $50\%$ as training data, $15\%$ as test data $test_{in}$ and $10\%$ as valiation data.
\begin{figure*}
    \centering
    \includegraphics[width=0.8\textwidth]{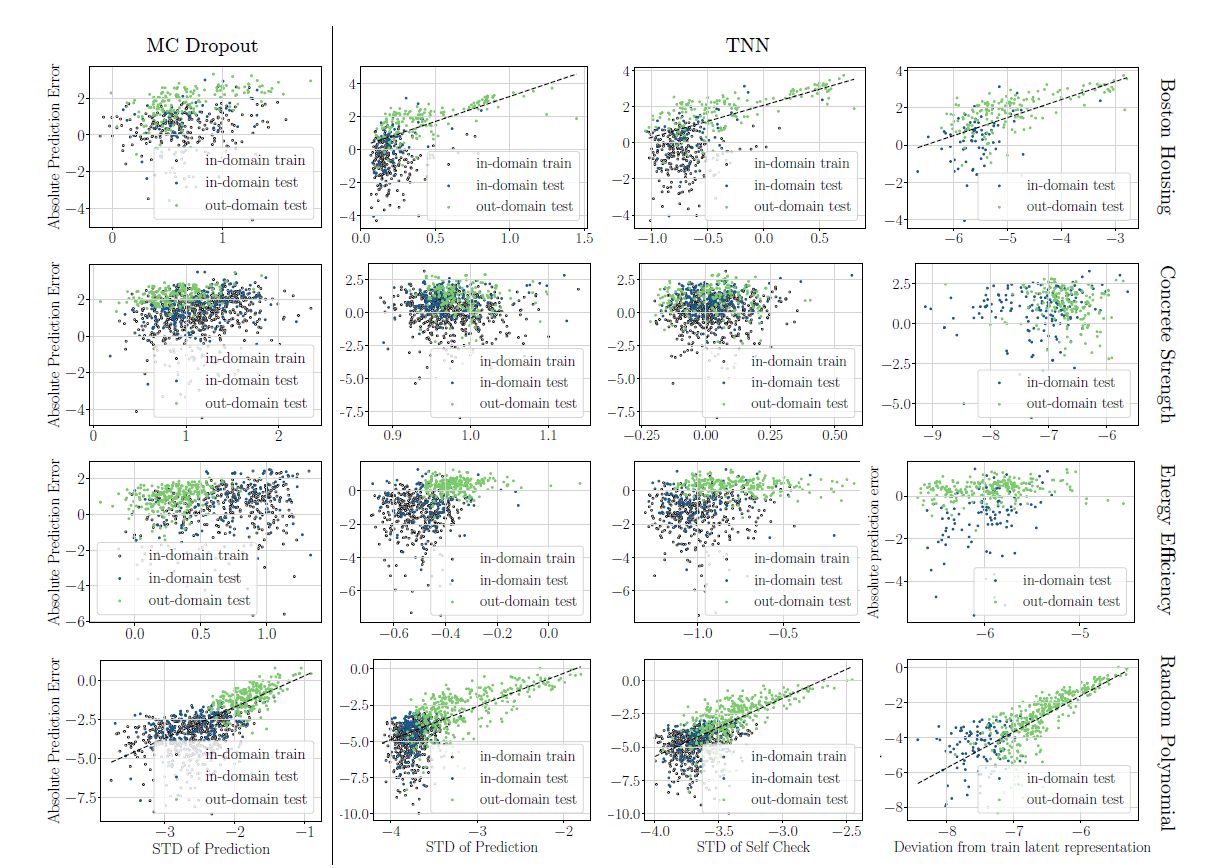}
    \caption{Comparison of different estimators for the prediction uncertainty, axis are logarithmic. Dropout at inference time is applied to single ANNs. For TNNs we examine the standard deviation of the prediction and the standard deviation of the self check consistency condition \eq{eq:sc2}. Further, we include the latent space distance to the training set. }
    \label{fig:manifold}
\end{figure*}
We examine if it is possible to estimate the prediction uncertainty using established methods. We perform Monte-Carlo dropout at an ANN \cite{gal2015dropout}. This method is based on applying dropout during the prediction phase to estimate the uncertainty of the prediction. In the case of TNNs we examine the standard deviations of the violations of the self-consistency conditions \eq{eq:sc2}, \eq{eq:sc3}, which includes the standard deviations of each single prediction. Finally, we compare the latent space distance \cite{Tran2020} to the training data of each prediction to its prediction uncertainty. In this case the projection into latent space is the output of the second last layer of the neural network.

The results of these examinations can be found in \fig{fig:manifold}. We differentiate between predictions on three different data sets: training set, in-domain test set $test_{in}$ and out-of-domain test set $test_{out}$. A general observation is that the prediction error of each single sample has the tendency to be the smallest on the training set, higher on $test_{in}$ and even higher on the $test_{out}$ (\tab{tab:manifold_results}). We can also see that on the training set not a single method for the prediction uncertainty estimation is accurate. However, we find a strong correlation between some of the methods and the prediction error on the test sets $test_{in}$ and $test_{out}$. There is no clear evidence that any of the prediction uncertainty estimation methods is uniformly better than any other, however, dropout seems to be worse than the other methods.

In our data sets, on the test sets there is empirical evidence that the prediction uncertainty can be modeled with functions RMSE$(\sigma)\approx a\times \sigma^\alpha$ and RMSE$(d) \approx b\times d^\beta$, where $\sigma$ is the standard deviation $d$ the latent space distance and $a,\alpha,b,\beta$ data set dependent parameters.

Further, we make a practical observation. As long as any of the predictors, applied to an unseen data point, is of the same magnitude as on the in-domain test set $test_{in}$ we expect the prediction uncertainty best estimated by RMSE$_{test_{in}}$. If any of the estimators is larger than that (as it is often the case on $test_{out}$), we expect the prediction error to be larger than the test error. This observation explains why the TNN self-consistency check can be employed as a proxy for decreasing prediction accuracy.

\section{Conclusions}
We have introduced twin neural network regression (TNN regression). It is based on the reformulation of a traditional regression problem into a prediction for the difference between two regression targets after which an ensemble is created by averaging the differences anchored at all training data points. This bears two advantages: a) the number of training data points increases quadratically over the original data set b) By anchoring each prediction at all training data points one obtains an ensemble of predictions of twice the training data set size.
Although a straightforward comparison is difficult, compared to trajectory ensemble methods \cite{huang2017snapshot}, which typically produce approximately $\times5$ snapshots per real training time equivalent, TNN regression produces on our data sets at least $\times 25$ ensemble members. We have demontrated that TNN regression can compete and outperform traditional regression methods including ANN regression, random forests and xgboost. Building an ensemble of TNNs can improve the predictive accuracy even further (\tab{tab:results}). TNNs are competitive with tree based methods on discrete datasets, however xgboost is significantly faster to train. On continuous data sets and on image based data sets TNNs are the clear winner. 
In the case where there is a large number of training data, TNNR might not see all possible pairs before convergence, such that it can't leverage its full advantage over traditional ANNs. In this case ANNs are able to compete with TNNR as shown in \fig{fig:data_rmse}. Since the anchor points during inference are only linear in the number of training data points sampling is normally not necessary. A successfully trained TNN must satisfy many self-consistency conditions. The violation of these conditions give rise to an uncertainty estimate of the final prediction (\fig{fig:manifold}). Future directions include intelligently weighting the ensemble members by improved ensembling techniques. Some problems might benefit from exchanging the ensemble mean by the median. It would also be interesting to examine the ensemble diversity of TNN regression compared to other ensembles. 





\bibliography{library}

\bibliographystyle{icml2021}
\newpage
\section{Supplementary Material}

\subsection{Datasets}
We describe the properties of the data sets in \tab{tab:data}. In addition to the common reference data sets we included scientific data and image like data for regression. The random polynomial function (RP) is a data set created from the equation
\begin{align}
F(x_1,...,x_5)=\sum_{i,j=0}^5a_{ij}x^{i}x^{j}+\sum_i^5 b_i x^{i} +c
\end{align}
with random but fixed coefficients and zero noise.

The output in the RCL circuit current data set (RCL) is the current through an RCL circuit, modeled by the equation
\begin{align}
I_0=V_0 \cos(\omega t)/\sqrt{R^2+(\omega L-1/(\omega C))^2}
\end{align}
with added Gaussian noise of mean 0 and standard deviation 0.1.

The output of the Wheatstone Bridge voltage (WSB) is the measured voltage given by the equation
\begin{align}
V=U(R_2/(R_1+R_2)-R_3/(R_2+R_3))
\end{align}
with added Gaussian noise of mean 0 and standard deviation 0.1.
\begin{table*}

  \centering
  \begin{tabular}{lllll}
  Name & Key & Size & Features & Type\\
  Boston Housing & BH &506 & 13&Discrete, Continuous\\
  Concrete Strength & CS &1030&8&Continuous\\
  Energy Efficiency & EF &768&8& Discrete, Continuous \\
  Yacht Hydrodynamics & YH &308&6&Discrete\\
  Red Wine Quality & WN &1599&11&Discrete, Continuous\\
  Bio Concentration & BC &779&14&Discrete, Continuous\\
  Random Polynomial Function & RP &1000&5&Continuous\\
  RCL Circuit Current &RCL&4000&6&Continuous\\
  Wheatstone Bridge Voltage &WSB&200&4&Continuous\\
  Ising Model &ISING &2000&400&Image, Discrete

  \end{tabular}
    \caption{Datasets
  }
  \label{tab:data}
\end{table*}

\newpage

\subsection{Hyperparameter Optimization}
In this section we have included additional results of experiments on our data sets Boston Housing (BH), concrete strength (CS), energy efficiency (EE) and random polynomial function (RP). We have varied the $L_2$ regularization. We further exchanged our main optimizer \emph{adadelta} by \emph{rmsprop}. Finally, we examined the checkpoint that saves the weights based on the best performance on the validation set. We find that \emph{adadelta} seems to give more consistent results. As long as the $L_2$ penalty is not too large, there is no clear evidence to favour a certain $L_2$ penalty over another. We further examined modifications to the architecture, dropout regularization, different learning rate schedules in preliminary experiments, which are not listed here, since none of these changes led to significant and uniform improvement. 

\begin{table*}

  \centering
  \begin{tabular}{lllllll}
Data  & ANN & ANN (E)  & MC Dropout & TNN  & TNN (E)\\
\toprule
& & &$L_2$=0\\
\toprule
BH & $2.544 \pm 0.168$ & $2.780 \pm 0.252$ & $3.375 \pm 0.321$ & $2.656 \pm 0.243$ & $2.286 \pm 0.155$\\
CS & $4.964 \pm 0.161$ & $4.595 \pm 0.205$ & $5.941 \pm 0.187$ & $3.758 \pm 0.214$ & $3.247 \pm 0.166$\\
EE & $0.819 \pm 0.035$ & $0.791 \pm 0.030$ & $2.893 \pm 0.133$ & $0.494 \pm 0.018$ & $0.452 \pm 0.020$\\
RP & $0.045 \pm 0.003$ & $0.031 \pm 0.002$ & $0.087 \pm 0.003$ & $0.022 \pm 0.001$ & $0.012 \pm 0.001$\\
\toprule
& & &$L_2$=1e-6\\
\toprule
BH & $2.477 \pm 0.163$ & $2.743 \pm 0.282$ & $3.388 \pm 0.193$ & $2.476 \pm 0.159$ & $2.347 \pm 0.135$\\
CS & $4.669 \pm 0.185$ & $4.714 \pm 0.161$ & $5.995 \pm 0.186$ & $3.751 \pm 0.263$ & $3.398 \pm 0.228$\\
EE & $0.968 \pm 0.044$ & $0.777 \pm 0.034$ & $2.973 \pm 0.092$ & $0.502 \pm 0.033$ & $0.452 \pm 0.022$\\
RP & $0.049 \pm 0.002$ & $0.028 \pm 0.002$ & $0.079 \pm 0.004$ & $0.020 \pm 0.001$ & $0.011 \pm 0.001$\\
\toprule
& & &$L_2$=1e-5\\
\toprule
BH & $2.845 \pm 0.177$ & $2.202 \pm 0.082$ & $3.126 \pm 0.236$ & $2.931 \pm 0.244$ & $2.239 \pm 0.214$\\
CS & $4.599 \pm 0.157$ & $4.403 \pm 0.198$ & $5.675 \pm 0.194$ & $3.715 \pm 0.318$ & $3.480 \pm 0.275$\\
 EE & $1.029 \pm 0.035$ & $0.799 \pm 0.034$ & $2.905 \pm 0.123$ & $0.500 \pm 0.019$ & $0.433 \pm 0.019$\\
 RP & $0.064 \pm 0.002$ & $0.032 \pm 0.003$ & $0.086 \pm 0.004$ & $0.022 \pm 0.001$ & $0.012 \pm 0.001$\\
\toprule
& & &$L_2$=1e-4\\
\toprule
 BH & $2.875 \pm 0.160$ & $2.750 \pm 0.295$ & $3.330 \pm 0.231$ & $2.933 \pm 0.187$ & $2.359 \pm 0.187$\\
CS & $5.178 \pm 0.155$ & $4.249 \pm 0.128$ & $5.730 \pm 0.175$ & $3.607 \pm 0.193$ & $3.047 \pm 0.176$\\
EE & $1.763 \pm 0.030$ & $0.853 \pm 0.034$ & $2.747 \pm 0.106$ & $0.502 \pm 0.027$ & $0.448 \pm 0.024$\\
RP & $0.107 \pm 0.002$ & $0.024 \pm 0.001$ & $0.097 \pm 0.004$ & $0.032 \pm 0.002$ & $0.019 \pm 0.001$\\
\toprule
& & &$L_2$=1e-3\\
\toprule
 BH & $4.902 \pm 0.103$ & $2.368 \pm 0.175$ & $2.722 \pm 0.110$ & $2.293 \pm 0.119$ & $2.471 \pm 0.189$\\
 CS & $8.001 \pm 0.133$ & $4.034 \pm 0.136$ & $6.102 \pm 0.145$ & $3.547 \pm 0.198$ & $3.537 \pm 0.168$\\
EE & $4.651 \pm 0.059$ & $0.778 \pm 0.032$ & $2.877 \pm 0.098$ & $0.481 \pm 0.022$ & $0.467 \pm 0.021$\\
 RP & $0.268 \pm 0.004$ & $0.055 \pm 0.005$ & $0.110 \pm 0.004$ & $0.056 \pm 0.003$ & $0.052 \pm 0.003$\\
\toprule
  \end{tabular}
    \caption{Root mean squared errors (RMSEs) of the validation sets for different datasets, machine learning approaches, and regularizations. (Optimizer: adadelta, with model checkpoint saving)}
\end{table*}

\begin{table*}

  \centering
  \begin{tabular}{lllllll}
Data  & ANN & ANN (E)  & MC Dropout & TNN  & TNN (E)\\
\toprule
& & &$L_2$=0\\
\toprule
 BH & $3.093 \pm 0.140$ & $3.432 \pm 0.317$ & $2.953 \pm 0.154$ & $2.552 \pm 0.104$ & $2.614 \pm 0.204$\\
CS & $5.373 \pm 0.168$ & $5.136 \pm 0.207$ & $6.067 \pm 0.209$ & $4.186 \pm 0.249$ & $3.882 \pm 0.218$\\
EE & $0.984 \pm 0.033$ & $0.894 \pm 0.040$ & $2.957 \pm 0.117$ & $0.524 \pm 0.022$ & $0.456 \pm 0.020$\\
 RP & $0.050 \pm 0.002$ & $0.032 \pm 0.002$ & $0.086 \pm 0.002$ & $0.022 \pm 0.001$ & $0.016 \pm 0.001$\\
\toprule
& & &$L_2$=1e-6\\
\toprule
BH & $3.170 \pm 0.222$ & $2.749 \pm 0.200$ & $3.226 \pm 0.206$ & $2.869 \pm 0.233$ & $2.554 \pm 0.134$\\
 CS & $5.133 \pm 0.172$ & $4.963 \pm 0.192$ & $6.293 \pm 0.179$ & $4.030 \pm 0.225$ & $4.025 \pm 0.258$\\
 EE & $1.125 \pm 0.059$ & $0.992 \pm 0.042$ & $3.057 \pm 0.120$ & $0.590 \pm 0.027$ & $0.468 \pm 0.020$\\
RP & $0.054 \pm 0.003$ & $0.034 \pm 0.002$ & $0.085 \pm 0.003$ & $0.020 \pm 0.001$ & $0.011 \pm 0.001$\\
\toprule
& & &$L_2$=1e-5\\
\toprule
 BH & $3.184 \pm 0.173$ & $2.996 \pm 0.286$ & $3.780 \pm 0.377$ & $3.315 \pm 0.229$ & $3.201 \pm 0.384$\\
 CS & $4.991 \pm 0.221$ & $4.712 \pm 0.176$ & $5.934 \pm 0.139$ & $3.880 \pm 0.276$ & $4.044 \pm 0.228$\\
 EE & $1.255 \pm 0.043$ & $0.906 \pm 0.037$ & $2.943 \pm 0.122$ & $0.570 \pm 0.026$ & $0.477 \pm 0.026$\\
RP & $0.067 \pm 0.002$ & $0.029 \pm 0.002$ & $0.092 \pm 0.007$ & $0.026 \pm 0.002$ & $0.014 \pm 0.001$\\
\toprule
& & &$L_2$=1e-4\\
\toprule
 BH & $3.559 \pm 0.180$ & $2.890 \pm 0.180$ & $3.352 \pm 0.292$ & $3.173 \pm 0.254$ & $2.740 \pm 0.147$\\
CS & $5.799 \pm 0.171$ & $4.924 \pm 0.183$ & $5.777 \pm 0.177$ & $3.878 \pm 0.205$ & $3.743 \pm 0.251$\\
 EE & $1.891 \pm 0.045$ & $0.923 \pm 0.036$ & $2.973 \pm 0.103$ & $0.581 \pm 0.029$ & $0.477 \pm 0.021$\\
RP & $0.110 \pm 0.001$ & $0.035 \pm 0.005$ & $0.088 \pm 0.004$ & $0.032 \pm 0.002$ & $0.023 \pm 0.002$\\
\toprule
& & &$L_2$=1e-3\\
\toprule
BH & $5.382 \pm 0.163$ & $2.861 \pm 0.198$ & $3.566 \pm 0.279$ & $3.039 \pm 0.294$ & $3.048 \pm 0.270$\\
 CS & $8.325 \pm 0.140$ & $4.964 \pm 0.220$ & $6.132 \pm 0.180$ & $3.953 \pm 0.220$ & $4.186 \pm 0.233$\\
 EE & $4.714 \pm 0.054$ & $1.083 \pm 0.071$ & $2.891 \pm 0.122$ & $0.613 \pm 0.024$ & $0.533 \pm 0.026$\\
 RP & $0.270 \pm 0.004$ & $0.066 \pm 0.004$ & $0.113 \pm 0.006$ & $0.061 \pm 0.003$ & $0.054 \pm 0.003$\\
\toprule
  \end{tabular}
   \caption{Root mean squared errors (RMSEs) of the test sets for different datasets, machine learning approaches, and regularizations. (Optimizer: adadelta, with model checkpoint saving)}
\end{table*}

\begin{table*}

  \centering
  \begin{tabular}{lllllll}
Data  & ANN & ANN (E)  & MC Dropout & TNN  & TNN (E)\\
\toprule
& & &$L_2$=0\\
\toprule
BH & $2.865 \pm 0.270$ & $3.151 \pm 0.302$ & $3.328 \pm 0.287$ & $2.724 \pm 0.225$ & $2.548 \pm 0.155$\\
 CS & $5.209 \pm 0.247$ & $5.021 \pm 0.237$ & $6.221 \pm 0.178$ & $4.582 \pm 0.180$ & $4.148 \pm 0.147$\\
EE & $1.199 \pm 0.083$ & $0.888 \pm 0.042$ & $3.057 \pm 0.116$ & $0.828 \pm 0.062$ & $0.711 \pm 0.034$\\
 RP & $0.049 \pm 0.004$ & $0.031 \pm 0.002$ & $0.087 \pm 0.005$ & $0.025 \pm 0.001$ & $0.016 \pm 0.001$\\
\toprule
& & &$L_2$=1e-6\\
\toprule
BH & $3.040 \pm 0.234$ & $3.027 \pm 0.292$ & $3.384 \pm 0.208$ & $2.601 \pm 0.196$ & $2.582 \pm 0.147$\\
 CS & $5.600 \pm 0.203$ & $5.442 \pm 0.203$ & $6.135 \pm 0.185$ & $4.559 \pm 0.212$ & $4.327 \pm 0.198$\\
 EE & $1.259 \pm 0.098$ & $0.944 \pm 0.055$ & $2.961 \pm 0.101$ & $0.792 \pm 0.057$ & $0.788 \pm 0.023$\\
 RP & $0.050 \pm 0.003$ & $0.026 \pm 0.001$ & $0.077 \pm 0.004$ & $0.024 \pm 0.001$ & $0.015 \pm 0.001$\\
\toprule
& & &$L_2$=1e-5\\
\toprule
BH & $2.903 \pm 0.157$ & $2.418 \pm 0.096$ & $3.109 \pm 0.279$ & $3.228 \pm 0.283$ & $2.427 \pm 0.233$\\
 CS & $5.523 \pm 0.202$ & $5.212 \pm 0.183$ & $5.973 \pm 0.176$ & $4.607 \pm 0.236$ & $4.344 \pm 0.235$\\
 EE & $1.210 \pm 0.044$ & $0.920 \pm 0.042$ & $2.846 \pm 0.146$ & $0.795 \pm 0.055$ & $0.644 \pm 0.030$\\
 RP & $0.061 \pm 0.002$ & $0.030 \pm 0.003$ & $0.079 \pm 0.004$ & $0.030 \pm 0.001$ & $0.016 \pm 0.001$\\
\toprule
& & &$L_2$=1e-4\\
\toprule
 BH & $3.019 \pm 0.081$ & $3.068 \pm 0.320$ & $3.319 \pm 0.210$ & $3.062 \pm 0.183$ & $2.543 \pm 0.215$\\
 CS & $6.027 \pm 0.194$ & $4.837 \pm 0.156$ & $5.956 \pm 0.162$ & $4.318 \pm 0.161$ & $3.888 \pm 0.152$\\
 EE & $2.003 \pm 0.068$ & $0.970 \pm 0.041$ & $2.785 \pm 0.098$ & $0.784 \pm 0.048$ & $0.660 \pm 0.034$\\
 RP & $0.102 \pm 0.003$ & $0.026 \pm 0.001$ & $0.094 \pm 0.007$ & $0.040 \pm 0.002$ & $0.023 \pm 0.001$\\
\toprule
& & &$L_2$=1e-3\\
\toprule
BH & $4.887 \pm 0.107$ & $2.654 \pm 0.192$ & $2.656 \pm 0.091$ & $2.447 \pm 0.124$ & $2.721 \pm 0.205$\\
CS & $8.354 \pm 0.144$ & $4.520 \pm 0.129$ & $6.204 \pm 0.146$ & $4.260 \pm 0.160$ & $4.360 \pm 0.185$\\
EE & $4.592 \pm 0.075$ & $0.874 \pm 0.035$ & $2.380 \pm 0.109$ & $0.705 \pm 0.038$ & $0.634 \pm 0.024$\\
 RP & $0.266 \pm 0.004$ & $0.068 \pm 0.007$ & $0.112 \pm 0.004$ & $0.066 \pm 0.004$ & $0.060 \pm 0.004$\\
\toprule
  \end{tabular}
  \caption{Root mean squared errors (RMSEs) of the validation sets for different datasets, machine learning approaches, and regularizations. (Optimizer: rmsprop, no model checkpoint saving)}
\end{table*}

\begin{table*}

  \centering
  \begin{tabular}{lllllll}
Data  & ANN & ANN (E)  & MC Dropout & TNN  & TNN (E)\\\toprule
& & & $L_2=\text{0}$\\
\toprule
BH & $3.291 \pm 0.229$ & $3.326 \pm 0.344$ & $2.820 \pm 0.169$ & $2.462 \pm 0.085$ & $2.593 \pm 0.193$\\
CS & $5.185 \pm 0.229$ & $5.225 \pm 0.227$ & $6.116 \pm 0.157$ & $4.748 \pm 0.189$ & $4.382 \pm 0.172$\\
EE & $1.145 \pm 0.090$ & $0.851 \pm 0.034$ & $3.060 \pm 0.104$ & $0.844 \pm 0.052$ & $0.680 \pm 0.032$\\
 RP & $0.051 \pm 0.002$ & $0.029 \pm 0.002$ & $0.084 \pm 0.004$ & $0.024 \pm 0.001$ & $0.019 \pm 0.002$\\
\toprule
& & & $L_2$=1e-6\\
\toprule
 BH & $2.930 \pm 0.182$ & $2.748 \pm 0.179$ & $3.146 \pm 0.180$ & $2.882 \pm 0.254$ & $2.502 \pm 0.117$\\
 CS & $5.765 \pm 0.213$ & $5.423 \pm 0.178$ & $6.275 \pm 0.194$ & $4.790 \pm 0.163$ & $4.469 \pm 0.208$\\
 EE & $1.288 \pm 0.121$ & $1.028 \pm 0.057$ & $2.984 \pm 0.106$ & $0.826 \pm 0.041$ & $0.713 \pm 0.026$\\
RP & $0.049 \pm 0.003$ & $0.030 \pm 0.002$ & $0.083 \pm 0.002$ & $0.023 \pm 0.001$ & $0.015 \pm 0.001$\\
\toprule
& & &$L_2$=1e-5\\
\toprule

 BH & $3.344 \pm 0.283$ & $2.911 \pm 0.257$ & $3.567 \pm 0.396$ & $2.983 \pm 0.173$ & $3.155 \pm 0.383$\\
 CS & $5.867 \pm 0.178$ & $5.105 \pm 0.131$ & $6.195 \pm 0.156$ & $4.449 \pm 0.173$ & $4.563 \pm 0.179$\\
  EE & $1.210 \pm 0.042$ & $0.940 \pm 0.040$ & $2.845 \pm 0.133$ & $0.815 \pm 0.057$ & $0.679 \pm 0.029$\\
 RP & $0.060 \pm 0.002$ & $0.024 \pm 0.002$ & $0.081 \pm 0.004$ & $0.028 \pm 0.001$ & $0.017 \pm 0.001$\\
\toprule
& & &$L_2$=1e-4\\
\toprule

  BH & $3.313 \pm 0.191$ & $2.625 \pm 0.145$ & $3.362 \pm 0.305$ & $3.199 \pm 0.227$ & $2.803 \pm 0.134$\\
  CS & $5.905 \pm 0.136$ & $5.141 \pm 0.186$ & $5.956 \pm 0.179$ & $4.518 \pm 0.190$ & $4.123 \pm 0.206$\\
 EE & $2.010 \pm 0.076$ & $0.891 \pm 0.032$ & $2.932 \pm 0.095$ & $0.762 \pm 0.030$ & $0.611 \pm 0.021$\\
 RP & $0.101 \pm 0.002$ & $0.033 \pm 0.004$ & $0.082 \pm 0.004$ & $0.039 \pm 0.002$ & $0.026 \pm 0.002$\\
\toprule
& & &$L_2$=1e-3\\
\toprule

  BH & $5.197 \pm 0.193$ & $2.835 \pm 0.185$ & $3.296 \pm 0.251$ & $2.975 \pm 0.278$ & $3.028 \pm 0.267$\\
 CS & $8.381 \pm 0.153$ & $5.128 \pm 0.213$ & $6.153 \pm 0.155$ & $4.611 \pm 0.205$ & $4.678 \pm 0.216$\\
EE & $4.659 \pm 0.065$ & $0.954 \pm 0.039$ & $2.360 \pm 0.108$ & $0.747 \pm 0.036$ & $0.723 \pm 0.032$\\
 RP & $0.267 \pm 0.003$ & $0.068 \pm 0.005$ & $0.114 \pm 0.006$ & $0.066 \pm 0.004$ & $0.059 \pm 0.004$\\
\toprule
  \end{tabular}
   \caption{Root mean squared errors (RMSEs) of the test sets for different datasets, machine learning approaches, and regularization. (Optimizer: rmsprop, no model checkpoint saving)}
\end{table*}

\end{document}